\documentclass[a4paper, 10pt, conference]{ieeeconf}      %
\usepackage{FG2025}

\usepackage{url}
\usepackage{graphicx}
\usepackage{subcaption}
\usepackage{amsmath}
\usepackage{multirow}
\usepackage{array}
\usepackage{tabularx}
\usepackage{calc}
\usepackage{makecell}
\usepackage{floatrow}
\FGfinalcopy

\IEEEoverridecommandlockouts                              %
\title{\LARGE \bf
A Siamese Network to Detect If Two Iris Images Are Monozygotic
}

\author{\parbox{16cm}{\centering
    {\large Yongle Yuan and Kevin W. Bowyer}\\
    {\normalsize
    Department of Computer Science and Engineering \\
    University of Notre Dame\\}}
}

\begin{document}

\ifFGfinal
\thispagestyle{empty}
\pagestyle{empty}
\else
\author{Anonymous FG2025 submission\\ Paper ID \FGPaperID \\}
\pagestyle{plain}
\fi
\maketitle

\begin{abstract}

This study presents the first automated classifier designed to determine whether a pair of iris images originates from monozygotic individuals, addressing a previously untackled problem in biometric recognition.
In Daugman-style iris recognition, the textures of the left and right irises of the same person are traditionally considered as being as different as the irises of two unrelated persons.
However, previous research indicates that humans can detect that two iris images are from different eyes of the same person, or eyes of monozygotic twins, with an accuracy of about 80\%.
In this work, we employ a Siamese network architecture and contrastive learning to categorize a pair of iris images as coming from monozygotic or non-monozygotic irises.
This could potentially be applied, for example, as a fast, noninvasive test to determine if twins are monozygotic or non-monozygotic.
We construct a dataset comprising both synthetic monozygotic pairs (images of different irises of the same individual) and natural monozygotic pairs (images of different images from persons who are identical twins), in addition to non-monozygotic pairs from unrelated individuals, ensuring a comprehensive evaluation of the model's capabilities.
To gain deeper insights into the learned representations, we train and analyze three variants of the model using (1) the original input images, (2) iris-only images (masking everything but the iris region), and (3) non-iris-only images (masking the iris region). This comparison reveals that both iris texture and surrounding ocular structure contain information useful for the model to classify the image pairs as monozygotic or non-monozygotic.
% The results demonstrate that models leveraging full eye-region information outperform those trained on iris-only data, emphasizing the nuanced interplay between iris and ocular characteristics.
Our approach achieves accuracy levels using the full iris image that exceed those previously reported for human classification of monozygotic iris pairs.\footnote{Code is available on \url{https://github.com/yongleyuan/mz-iris-pairing}}

\end{abstract}

\section{INTRODUCTION}

\begin{figure}
    \centering
    \begin{subfigure}{0.48\textwidth}
        \centering
        \includegraphics[width=\linewidth]{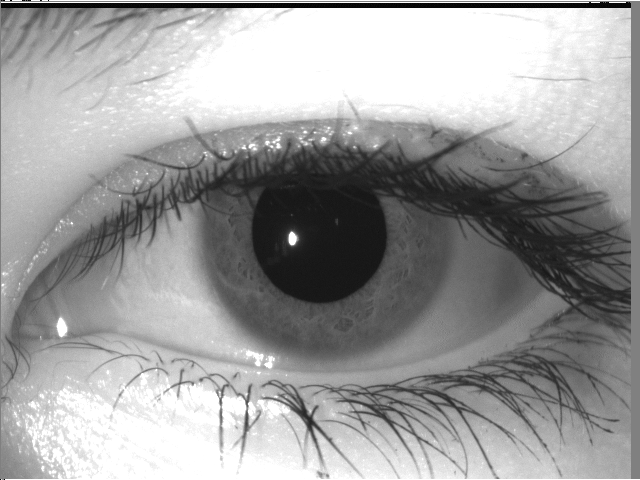}
        \caption{Pair 1 left iris}
        \label{fig:teaser_1_l}
    \end{subfigure}
    \hfill
    \begin{subfigure}{0.48\textwidth}
        \centering
        \includegraphics[width=\linewidth]{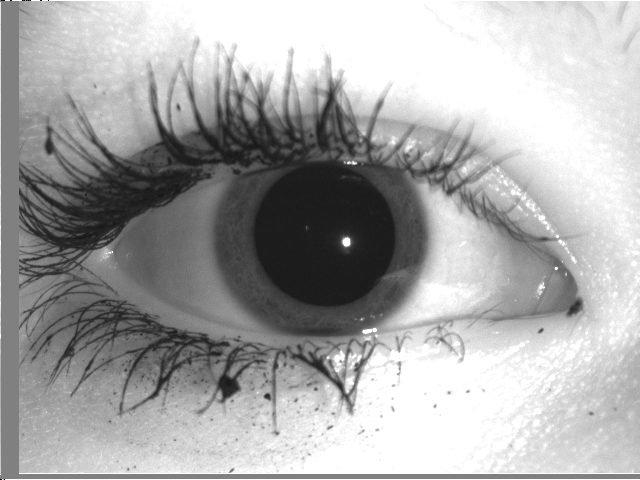}
        \caption{Pair 1 right iris}
        \label{fig:teaser_1_r}
    \end{subfigure}
    \begin{subfigure}{0.48\textwidth}
        \centering
        \centering
        \includegraphics[width=\linewidth]{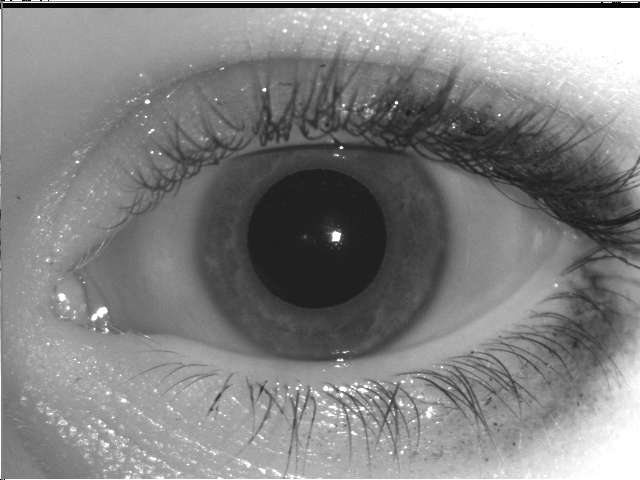}
        \caption{Pair 2 left iris}
        \label{fig:teaser_2_l}
    \end{subfigure}
    \hfill
    \begin{subfigure}{0.48\textwidth}
        \centering
        \centering
        \includegraphics[width=\linewidth]{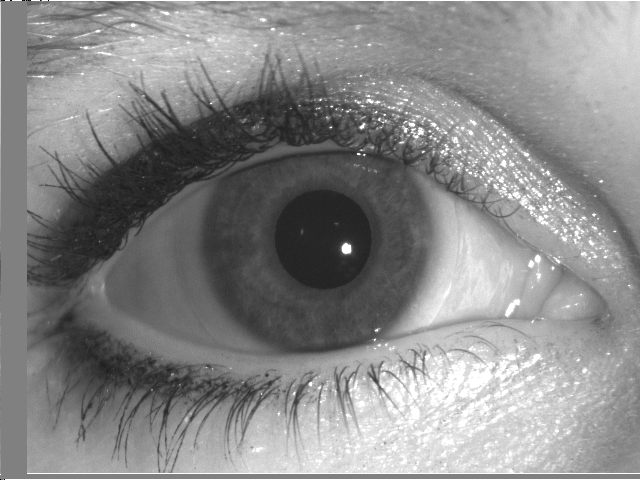}
        \caption{Pair 2 right iris}
        \label{fig:teaser_2_r}
    \end{subfigure}
    \caption{Monozygotic or not? Each of the four iris images is from a different person. One of the two pairs is from monozygotic twins. Can you tell which pair is monozygotic and which is not? The answer is at the end of the Acknowledgments section (Sec. \ref{sec:ack}).}
    \label{fig:teaser}
\end{figure}

Iris recognition has emerged as a leading biometric identification technology due to the unique and stable patterns present in the iris \cite{daugman2009iris}. These intricate patterns are formed randomly during fetal development \cite{wildes1997iris}. Iris recognition technologies are able to identify the distinctiveness between irises from different individuals. High accuracy, combined with speed and non-intrusive nature, makes iris recognition ideal for applications in security, border control, and identity verification. For example, India has used iris as one of the biometrics in the unique identification system called ``Aadhaar'' \cite{india2010aadhaar}.

While the principle that ``no two irises are the same'' supports the effectiveness of iris recognition technology \cite{daugman2009iris}, a notable limitation arises when dealing with monozygotic twins. Humans can often perceive subtle similarities in the iris patterns of monozygotic (``identical'') twins, yet iris biometric systems generally fail to detect these nuances \cite{hollingsworth2011genetically}. Monozygotic twins share identical genetic material due to originating from a single fertilized egg, leading to similarities in many visual characteristics, including iris patterns \cite{Larsson_Pedersen_Stattin_2003}. To our knowledge, no automated classifier has been previously developed to specifically distinguish between monozygotic and non-monozygotic iris pairs.

The ability to accurately determine if two individuals are monozygotic twins using iris recognition is crucial for several reasons. First, it offers a cost-effective and non-invasive alternative for determining if twins are identical or fraternal. Second, investigating how iris recognition systems detect — or fail to detect — subtle texture similarities contributes to explainable artificial intelligence (AI), improving our understanding of the decision-making processes of these systems and increasing their trustworthiness.

% \subsection*{Problem Definition}

The objective of this research is to develop a machine learning model capable of distinguishing between monozygotic and non-monozygotic twins. The contributions of this work include:

\begin{itemize}
    \item Assembling a dataset comprising monozygotic (MZ) and non-monozygotic (NMZ) pairs of iris images.
    \item Designing an experiment for identifying the two classes.
    \item Implementing a neural network model tailored for this task.
    \item Evaluating and analyzing the performance of the model to identify areas for improvement.
\end{itemize}

\section{RELATED WORKS}

Iris recognition has been extensively studied over the past several decades, stemming from the foundational work of Daugman \cite{daugman1993high}, who introduced the first commercially viable iris recognition system. His seminal approach represented the iris texture using a phase-coded representation derived from Gabor wavelets, and the resulting IrisCode allows for rapid and accurate matching. Daugman's system demonstrated that the iris is an extremely distinctive and stable biometric, capable of achieving high recognition accuracy with a very low false match rate.

While iris recognition has matured as a technology \cite{daugman2007new, nguyen2024deep, patil2016survey}, research focusing on MZ twins remains relatively limited. Traditional iris recognition systems rely heavily on the assumption that inter-individual iris differences are sufficiently pronounced that false matches are improbable, an assumption largely validated by early works \cite{daugman1993high}. However, Hollingsworth found that humans can perceive a level of similarity in the iris patterns in MZ twins that is not ``seen'' by handcrafted texture features, including those derived from the original IrisCode framework or subsequent wavelet-based representations \cite{hollingsworth2011genetically}.  Other studies compared the distribution of different-iris similarity scores for MZ twins and for unrelated persons, and reported finding no significant difference between the distributions \cite{daugman2001epigenetic, matey2020analysis}.

In recent years, deep learning techniques have gained traction in the iris recognition domain. Such methods automatically learn details that may not be readily represented by handcrafted descriptors \cite{gangwar2016deepirisnet}. Siamese networks, in particular, have been applied to various biometric tasks including face \cite{chopra2005learning}, palmprint \cite{zhong2018palmprint}, and fingerprint \cite{zhu2021fingerprint} recognition, providing a powerful framework for learning similarity metrics directly from data. Gautam et al. \cite{gautam2021identifying} have extended this approach by exploring the use of Siamese architecture for twin identification based on ocular region features, including the iris and surrounding areas. Their work demonstrates that fusing multi-scale information from the entire ocular region can achieve human-level accuracy. Despite their success, application of Siamese networks to MZ twin iris recognition remains underexplored.

This study builds on these foundations by leveraging a Siamese architecture and contrastive learning to identify MZ twins. By doing so, we seek to bridge the gap between early, highly effective statistical approaches and contemporary deep learning methods, advancing the state-of-the-art in this challenging sub-problem of iris recognition. To our knowledge this is the first work to develop a classifier to tell if a pair of iris images is from MZ persons or not.

\section{DATASET}

This section describes the datasets used in our research and the methodology for constructing the training, validation, and test sets.

\subsection{ISO/IEC 29794-6 Quality Metrics}
\label{sec:iso-metrics}

Poor-quality images can introduce noise and artifacts that negatively impact training. To screen out poor-quality images, we use the ISO/IEC 29794-6 standard for iris image quality metrics \cite{isometric}. This standard provides a comprehensive framework for evaluating iris image quality using the factors described in Appendix \ref{appx:iso-metrics}.

\subsection{Monozygotic Iris Pairs}

\begin{figure*}[]
    \centering
    \includegraphics[width=\textwidth]{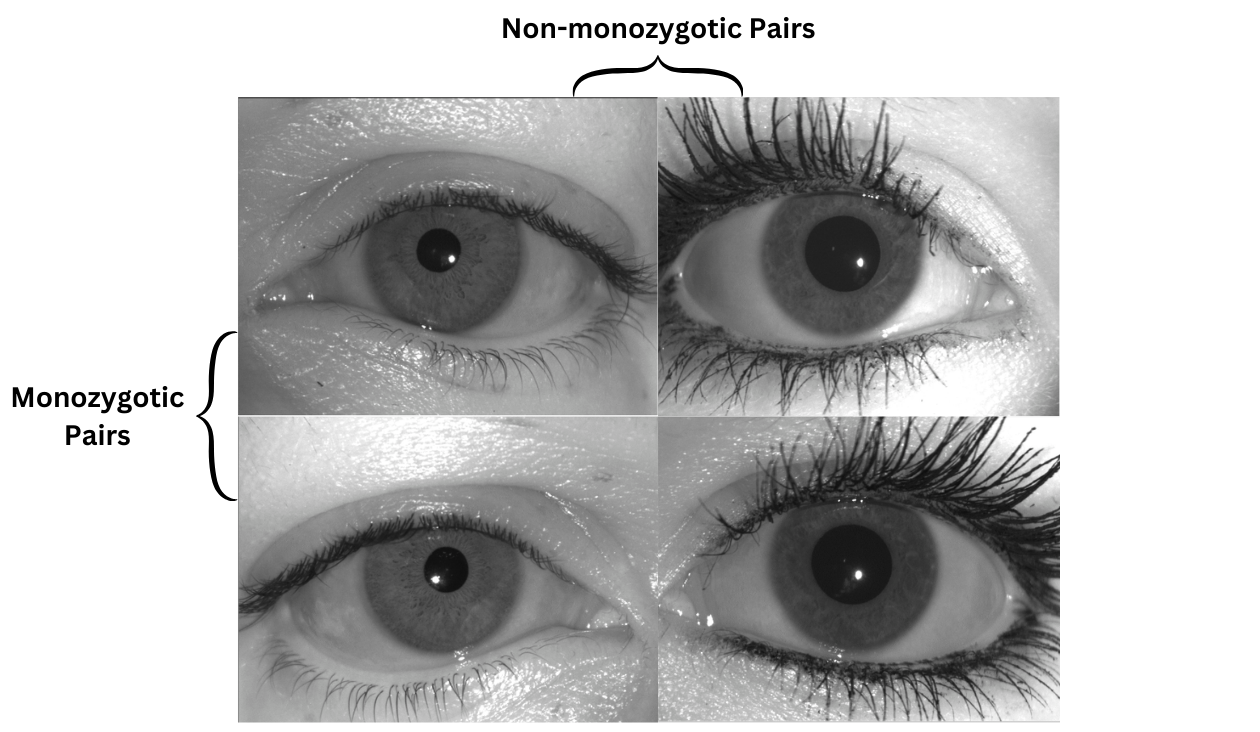}
    \caption{Comparisons of MZ and NMZ pairs. Vertical image pairs are MZ pairs, while horizontal image pairs are NMZ pairs. The left MZ pair features well-defined, strip-like textures near the pupil, while the right MZ pair has lighter, polygon-like lines. Note: (1) the two irises in each MZ pair have similar dilation, but this is not necessarily the case throughout the dataset and (2) these MZ pairs are \textit{synthetic}.}
    \label{fig:mz_vs_nmz}
\end{figure*}

MZ iris pairs refer to iris images that originate from genetically identical sources. 
Images from the left and right eye of the same person are an example of MZ irises.
Images from the left and right iris of each person in a pair of identical twins are another example of a group of MZ irises.
We refer to iris image pairs from the same person as \textit{synthetic MZ pairs}, and those from actual identical twins as \textit{natural MZ pairs}. Fig. \ref{fig:mz_vs_nmz} shows a comparison between MZ and NMZ pairs. In our study, we collect MZ and NMZ pairs from sources described below.

\subsection{Data Sources}

All images are grayscale acquired in near-infrared illumination with resolution of 640×480 pixels, sourced from the following two datasets. These images significantly overlap with those used in Hollingsworth's research investigating human ability to distinguish MZ and NMZ pairs \cite{hollingsworth2011genetically}. A comparative analysis of their findings and our results is in Sec. \ref{sec:comp-human-perf}.

\subsubsection{ND-IRIS-0405}

The ND-IRIS-0405 dataset, collected by the University of Notre Dame’s Computer Vision Research Lab \cite{nguyen2024deep, phillips2009frvt}, contains 64,980 iris images from 356 subjects. Images include left and right eyes taken under controlled conditions across multiple sessions, ensuring diversity for robust model training.

\subsubsection{BXGrid}

The BXGrid is a large-scale, closed-source repository for biometric research at the University of Notre Dame \cite{bui2009experience}. It has over half a million iris images, including those in the ND-IRIS-0405 dataset. It also includes iris images from identical twins. These images are used to evaluate the performance of our model's ability in real-world scenarios.

\subsection{Dataset Construction} \label{sec:dataset-construction}

% We would like to use MZ twins for both training and testing. However, due to an insufficient number of available MZ twin samples, we use synthetic MZ iris pairs for training and actual MZ twin iris data for testing. This setup allows us to train the model on genetically identical irises while capturing intra-subject variations. However, we acknowledge that subtle differences might exist between synthetic and natural MZ iris pairs. This is a potential area for future investigation with larger natural MZ datasets.

We know of no large datasets of MZ twin iris images available to the research community. This is a challenge for developing a training set for creating a model. However, the left and right irises of the same person can serve as a proxy for MZ irises. Therefore, we create a large training set of MZ iris image pairs by using left and right irises of the same person as a substitute for MZ iris image pairs coming from  different persons of a MZ set of twins. The similarities and differences of these two types is a topic that merits further study when such data becomes more available. Nevertheless, results in this paper show that this approach to assembling a training set is effective.

The training and validation set are synthesized from the ND-IRIS-0405 dataset. Positive samples consist of left and right iris images from the same individual taken on different days (synthetic MZ pairs). This setup aims to capture intra-subject variations while maintaining the inherent biometric similarities for MZ irises. The same number of negative samples is created by pairing left and right iris images from different individuals to present inter-subject variations. To avoid low-quality images in training and validation datasets, we screen out images with an overall ISO quality score below a threshold of 50 and those with scores of 255, meaning computation failure.

The test set is derived from the twins' iris images in BXGrid repository. These image pairs mimic real-life scenarios when identifying if two people are identical or not (natural MZ pairs). Positive pairs are formed by pairing the left iris image of one twin with the right iris image of the other twin. This setup challenges the model to use what is learned from the synthetic MZ pairs on natural MZ pairs. By avoiding mixing them, we minimize any bias that the model might have towards either the synthetic or the natural MZ pairs. It is important to note that the MZ twin pairs in this test set are based on self-identification. No genetic or clinical tests such as DNA or blood tests were conducted to confirm zygosity. It is possible that some of the pairs presumed to be monozygotic may not be genetically identical.

The assembled training dataset is balanced, consisting of 23,814 image pairs (11,907 positive pairs and 11,907 negative pairs) from 63 subjects of various races, including White, Asian, and Hispanic. Among these 63 subjects, 34 are female, and 29 are male. The test dataset is also balanced, with 824 positive pairs from 193 individuals and 824 negative pairs from 191 individuals.

The training and validation data used in our work is available to the research community on the website of the Computer Vision Research Lab at Notre Dame\footnote{https://cvrl.nd.edu/projects/data/}.  We do not have the rights to re-distribute the MZ twins iris images \cite{matey2020analysis}, but the list of image pairs used will be made available.

\section{METHODOLOGY}

The primary objective of this study is to distinguish between MZ and NMZ iris pairs. Instead of framing this as a traditional classification task with discrete classes, we adopt a contrastive learning approach. Contrastive learning aims to learn encoded representations where similar samples are pulled closer together, and dissimilar samples are pushed farther apart in the embedding space \cite{khosla2020contrastivelearning}. By minimizing the Euclidean distance between embeddings of positive (MZ) pairs and maximizing it for negative (NMZ) pairs, the model learns to capture the subtle differences in iris patterns that are critical for accurate discrimination.

To implement the contrastive learning strategy, we utilize a Siamese network architecture \cite{chicco2021siamese}. While this architecture is established for similarity learning tasks, it is applied to this specific iris biometric challenge for the first time to our knowledge. A Siamese network consists of two identical sub-networks with shared weights and parameters, ensuring both inputs are processed in the same manner. Each sub-network functions as an encoder to map an input image to a fixed-dimensional embedding in the vector space. The similarity between the two embeddings is then evaluated using a distance metric, such as Euclidean distance. The Siamese network is particularly well-suited for similarity-based tasks, as it enforces consistency in the representation of both inputs and enables the network to focus on learning meaningful differences between MZ and NMZ pairs.

\section{EXPERIMENT}

\subsection{Data Preprocessing}

\begin{figure}[t]
    \centering
    \begin{subfigure}{0.48\textwidth}
        \centering
        \includegraphics[width=\linewidth]{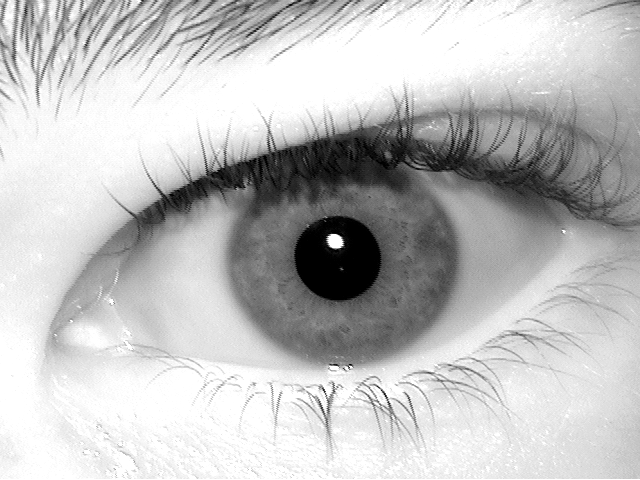}
        \caption{Raw input image\\(left iris)}
        \label{fig:mask-example-a}
    \end{subfigure}
    \hfill
    \begin{subfigure}{0.48\textwidth}
        \centering
        \includegraphics[width=\linewidth]{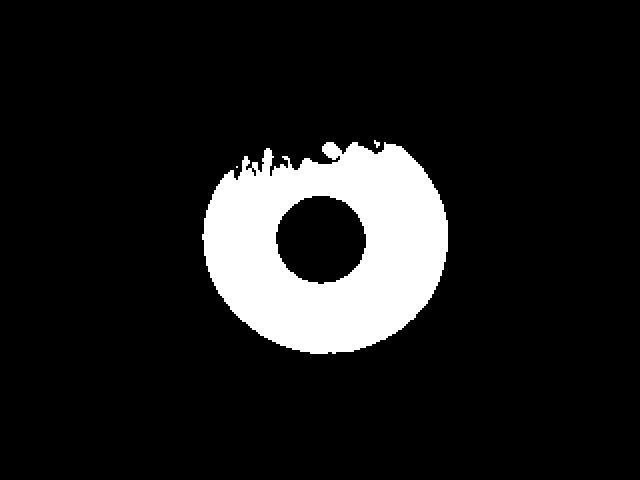}
        \caption{Seg. mask\\(left iris)}
        \label{fig:mask-example-b}
    \end{subfigure}
    \hfill \\
    \begin{subfigure}{0.48\textwidth}
        \centering
        \includegraphics[width=\linewidth]{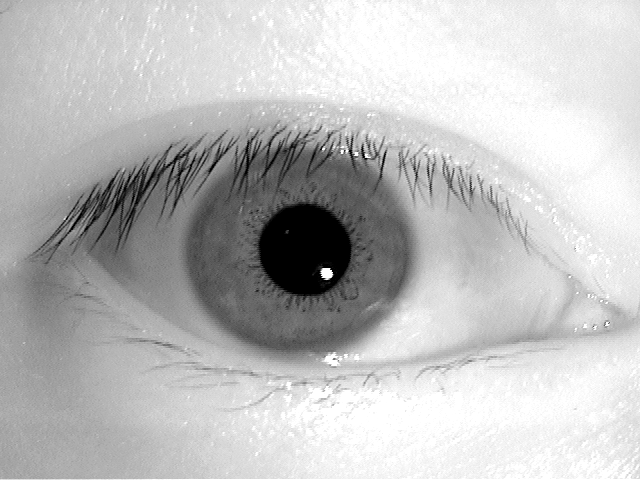}
        \caption{Raw input image\\(right iris)}
        \label{fig:mask-example-c}
    \end{subfigure}
    \hfill
    \begin{subfigure}{0.48\textwidth}
        \centering
        \includegraphics[width=\linewidth]{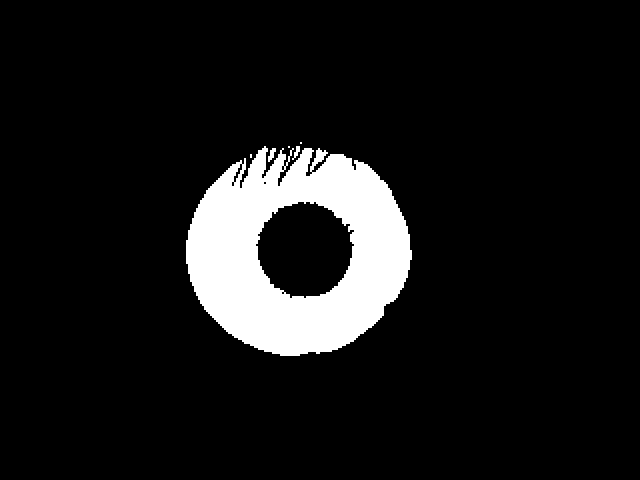}
        \caption{Seg. mask\\(right iris)}
        \label{fig:mask-example-d}
    \end{subfigure}
    \hfill \\
    \caption{Raw iris images and corresponding segmentation masks.}
    \label{fig:mask-example}
\end{figure}

The input images can be analyzed in three distinct ways, based on the area of the image contributing to the decision.
\begin{enumerate}
    \item Raw images: Original iris sensor images with a resolution of 640 * 480, including the full eye region (pupil, iris, sclera, eyelid, eyelashes, and eyebrows).
    \item Iris-only images (masked images): Non-iris regions are removed using pixel-wise segmentation masks, isolating the iris region (see Figures \ref{fig:mask-example-b} and \ref{fig:mask-example-d}).
    \item Non-iris-only (inversely masked images): Only non-iris regions are retained as a control to identify which regions influence the model’s decisions.
\end{enumerate}

\begin{figure}[t]
    \centering
    \begin{subfigure}{0.48\textwidth}
        \centering
        \includegraphics[width=\linewidth]{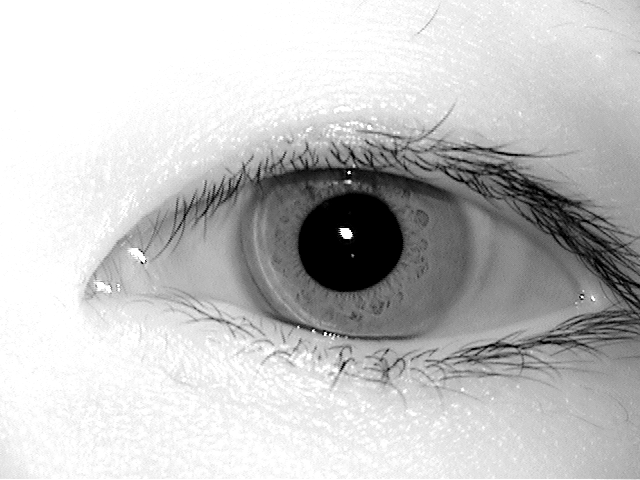}
        \caption{Original}
        \label{fig:img_orig}
    \end{subfigure}
    \hfill \\
    \hfill \\
    \begin{subfigure}{0.48\textwidth}
        \centering
        \includegraphics[width=\linewidth]{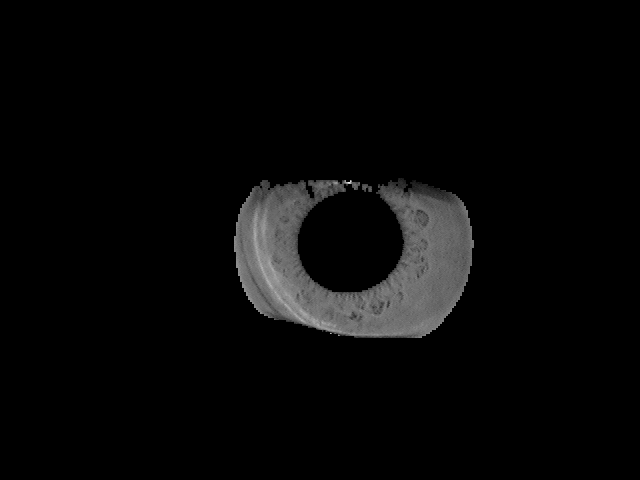}
        \caption{Iris-only}
        \label{fig:img_mask}
    \end{subfigure}
    \begin{subfigure}{0.48\textwidth}
        \centering
        \includegraphics[width=\linewidth]{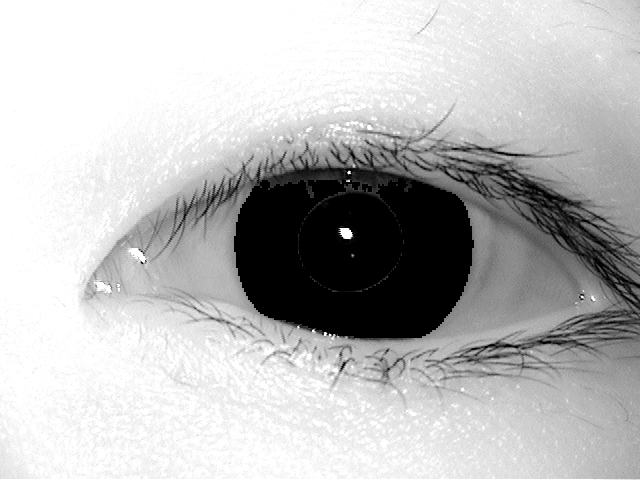}
        \caption{Non-iris-only}
        \label{fig:img_mask_inv}
    \end{subfigure}
    \caption{One example of original input images (\ref{fig:img_orig}) with its corresponding iris-only (\ref{fig:img_mask}) and non-iris-only input (\ref{fig:img_mask_inv}).}
    \label{fig:input_example}
\end{figure}

Segmentation masks were generated using the encoder-decoder network of Trokielewicz et al. \cite{rokielewicz2020post}, ensuring accurate filtering of non-iris regions. Examples of raw images and their segmentation masks are shown in Fig. \ref{fig:mask-example}. Fig. \ref{fig:input_example} shows examples of an input image with its corresponding iris-only and non-iris-only versions.

\subsection{Model Backbone}

We employ ResNet-18 as the backbone for our image encoder due to its proven effectiveness in various computer vision tasks \cite{he2016resnet}. ResNet-18 is a convolutional neural network (CNN) known for its residual connection, which addresses the vanishing gradient problem by allowing gradients to flow through shortcut connections that skip one or more layers. ResNet has multiple common variants, of which ResNet-18 is the shallowest. It is effective and easy to train.  
We explored deeper ResNet backbones (34, 50, 101) but found that they did not offer any clear increase in accuracy for this problem; see Appendix \ref{appx:abl-backbone-depth}.

To adapt ResNet-18 for our specific task, we modify the final layer of the network. The original model is designed and trained for classification tasks on the ImageNet dataset and includes a final fully connected layer with 1,000 neurons corresponding to 1,000 classes \cite{deng2009imagenet, he2016resnet}. We replace this layer with a new fully connected layer comprising 128 neurons. This change reduces the dimensionality of the output from 1,000 to 128, creating a compact embedding that serves as image embeddings of the input iris images.

Each input image in a pair is passed through the modified ResNet-18 encoder, resulting in two 128-dimensional embedding vectors. These embeddings capture the salient features from the input image, encoding the fine-grained yet unique structural and textural information necessary for distinguishing between MZ and NMZ pairs.

\subsection{Experiment Setup}

\subsubsection{Distance Metric}

The similarity between embeddings is measured using the Euclidean distance:
\begin{equation}
    D(\mathbf{e}_1, \mathbf{e}_2) = \sqrt{||\mathbf{e}_1 - \mathbf{e}_2||_2}
    \label{eq:distance}
\end{equation}
where $\mathbf{e}_1$ and $\mathbf{e}_1$ are the embeddings of the two images in a pair. A smaller distance indicates higher similarity.

\subsubsection{Loss Function}
\label{sec:loss}

Training a model typically requires a loss function, which is used during backpropagation \cite{werbos1990backpropagation}. We use the contrastive loss function:
\begin{equation}
    \mathcal{L} = \frac{1}{2} \left( y_\text{pred} \cdot D^2 + (1-y_\text{pred}) \cdot \max(0, m-D)^2 \right)
    \label{eq:loss}
\end{equation}
where $\mathcal{L}$ is the loss, $y_\text{pred}$ is a binary label indicating whether the pair is similar/positive/MZ ($y_\text{pred}=1$) or dissimilar/negative/NMZ ($y_\text{pred}=0$), $D$ is the Euclidean distance in (\ref{eq:loss}), and $m$ is the margin that defines the maximum acceptable distance between negative pairs. We set $m=1$ to normalize the distance between 0 and 1 after training so that we can easily recognize outliers if there are any. Since the distance should be normalized, we intuitively set the distance decision boundary/threshold to the midpoint: 0.5. We will discuss how this threshold affects the training and testing in Appendix \ref{appx:abl-dist-thres}.

When $y_\text{pred}=1$, the right part of the addition in (\ref{eq:loss}) is $0$, leaving only the left part, which should be small if the pairs are close in the vector space (correct prediction) and large if the pairs are far from each other in the vector space (incorrect prediction). When $y_\text{pred}=0$, only the right part of the addition is left. The loss should then be closer to $m$ if the prediction is correct and vice versa. Averaging the two parts ensures that the loss is large for falsely predicted pairs and small for correct predictions, setting the direction for gradients during backpropagation.

\subsubsection{Optimizer}
\label{sec:optimizer}

To train the proposed model effectively, we employ the Adam optimizer, a stochastic, gradient-based adaptive moment estimation algorithm designed for training deep neural networks \cite{kingma2017adam}. 
We set the learning rate as 0.0001, while the rest of the parameters are kept as their default values ($\beta_1=0.9$, $\beta_2=0.999$, $\epsilon=1e-7$).

\subsubsection{Evaluation Metrics}

To assess the model's performance, we use the following evaluation metrics.
\begin{itemize}
    \item Accuracy: The proportion of correctly identified pairs over the total number of pairs.
    \item Precision: The proportion of correctly identified positive pairs over all positive predictions.
    \item Recall: The proportion of correctly identified positive pairs over all true positive pairs.
    \item Distance: Distance between predicted positive and negative pairs.
\end{itemize}

\subsection{Model Training}

The model is trained separately using three different types of inputs: (1) original iris images, (2) iris-only images, and (3) non-iris-only images. Comparing iris-only and non-iris-only results allows us to analyze how the parts of the image contribute to the task.
% The non-iris-only images function as a control group to provide insights on what areas the model is focusing on.

30\% of the training data are randomly selected as validation data. These image pairs are not inputted in any forward passes. Instead, they are used after each forward pass to compute the evaluation metrics, which will be compared with those collected after inference.

We initialized the ResNet-18 backbone with the ImageNet pre-trained weights, which should already have some prior knowledge in distinguishing between different objects. As is standard for Siamese networks, we then fine-tuned the model with weights being shared across two identical backbones. Each time the model is trained with a batch size of 32 pairs. The loss function and the optimizer, as mentioned in Sec. \ref{sec:loss} and \ref{sec:optimizer}, are kept the same throughout. All experiments are run 5 times, with each individual run having the same train-validation split and data batching.

\subsection{Training Results}

\begin{figure}
    \centering
    \includegraphics[width=\linewidth]{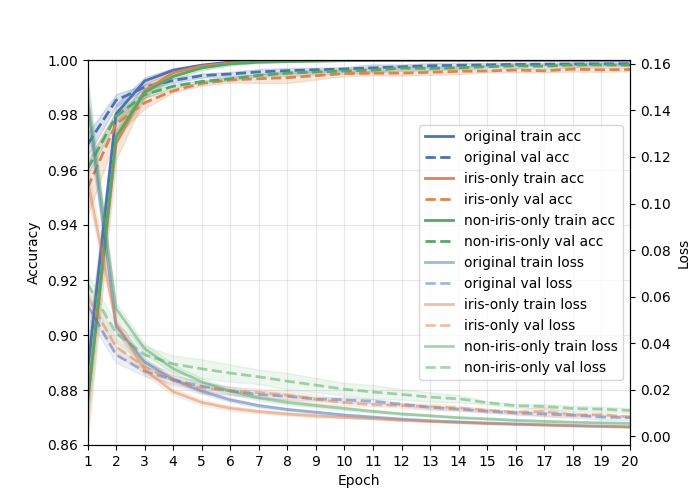}
    \caption{Training and validation accuracy and loss over epochs. The data points are the average value over 5 runs. The shaded areas represent $\pm1$ standard deviation of the 5 runs.}
    \label{fig:train-val-graph}
\end{figure}

Across all three types of input images, the model demonstrated rapid convergence during training. Fig. \ref{fig:train-val-graph} illustrates accuracy and loss during training. The training accuracy for all three input types quickly approaches 1, indicating that the model can distinguish between MZ and NMZ pairs in nearly all training samples after epoch 7. The minimal separation of lines suggests that the input type has a negligible impact on final training accuracy. The training loss follows a similar pattern of convergence, reaching near-zero around epoch 7. The iris-only image pairs show a faster decline in loss, suggesting that the iris information makes the model converge faster and easier under the same setting, compared to analyzing complete iris images and images without iris texture.

The validation results closely mirror the trends observed in the training phase, but they have some fluctuations as they converge as shown in Fig. \ref{fig:train-val-graph}. Validation accuracy for all three input types approaches 1. The validation loss decreases over time smoothly but is slightly higher than the training loss, indicating sensitivity to the variability within the validation data.

We collected the distance between positive and negative pairs. All three input types' positive and negative distances rapidly converge to their optimal values (0 and 1 respectively) in both training and validation stages.

\subsection{Testing Results}

After training our models, we evaluate them on the test set derived from BXGrid. This test set includes genuine MZ twin pairs, enabling an assessment of the model's generalizability. We test all five runs of the three models trained with original input, iris-only input, and non-iris-only input, using the best epoch (with the lowest loss). The final results are obtained by averaging these five runs.

\begin{table*}[h!]
\centering
\begin{tabular}{cccccccccc}
\Xhline{1pt}
\textbf{Model Input} & \textbf{Accuracy} & \textbf{Precision} & \textbf{Recall} & \textbf{F1 Score} & \textbf{Avg Pos Dist} & \textbf{Avg Neg Dist} \\
\hline
Original & $\mathbf{0.81}\pm0.024$ & $0.82\pm0.037$ & $\mathbf{0.81}\pm0.078$ & $\mathbf{0.81}\pm0.034$ & $0.38\pm0.038$ & $0.77\pm0.086$ \\
Iris-only & $0.73\pm0.017$ & $0.80\pm0.031$ & $0.60\pm0.060$ & $0.69\pm0.033$ & $0.47\pm0.028$ & $0.77\pm0.077$  \\
Non-iris-only & $0.77\pm0.038$ & $\mathbf{0.84}\pm0.034$ & $0.68\pm0.100$ & $0.74\pm0.063$ & $0.45\pm0.049$ & $0.87\pm0.097$  \\
\Xhline{1pt}
\end{tabular}
\caption{Testing results. The reported values are obtained by averaging the testing results over 5 experiment runs. All values are coupled with their standard deviation (after $\pm$). Higher values represent better performance. \textbf{Bold values} are best results among three model inputs.}
\label{tab:test-res}
\end{table*}

\begin{figure}[]
    \centering
    \begin{subfigure}{0.48\textwidth}
        \centering
        \includegraphics[width=\linewidth]{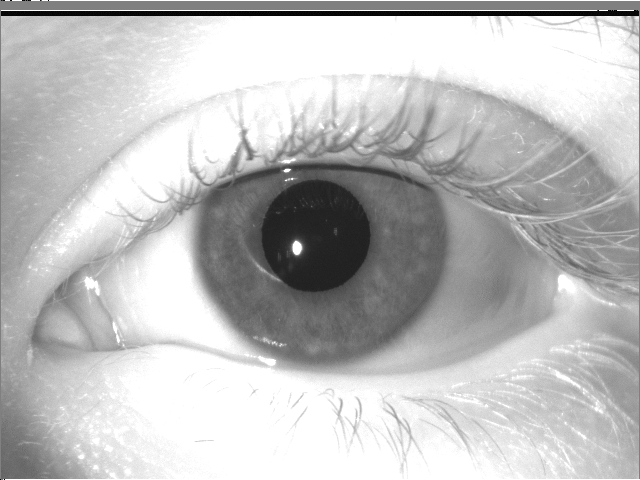}
        \caption{FP example (left)}
        \label{fig:test_example_fp_l}
    \end{subfigure}
    \hfill
    \begin{subfigure}{0.48\textwidth}
        \centering
        \includegraphics[width=\linewidth]{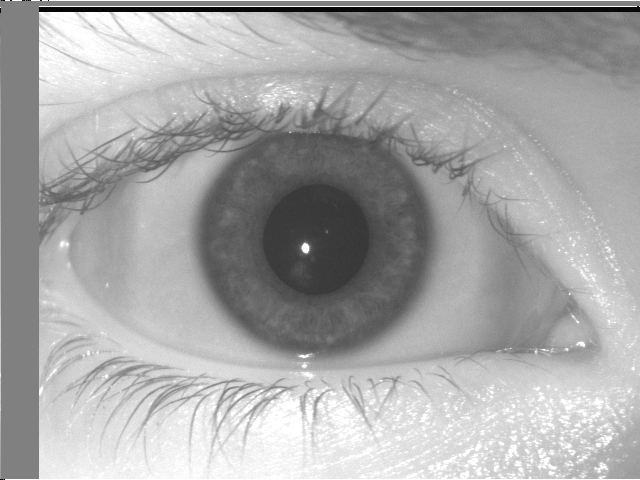}
        \caption{FP example (right)}
        \label{fig:test_example_fp_r}
    \end{subfigure}
    \begin{subfigure}{0.48\textwidth}
        \centering
        \centering
        \includegraphics[width=\linewidth]{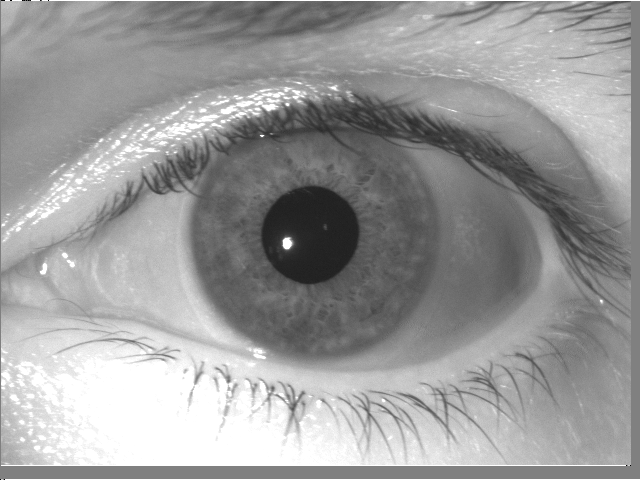}
        \caption{FN example (left)}
        \label{fig:test_example_fn_l}
    \end{subfigure}
    \hfill
    \begin{subfigure}{0.48\textwidth}
        \centering
        \centering
        \includegraphics[width=\linewidth]{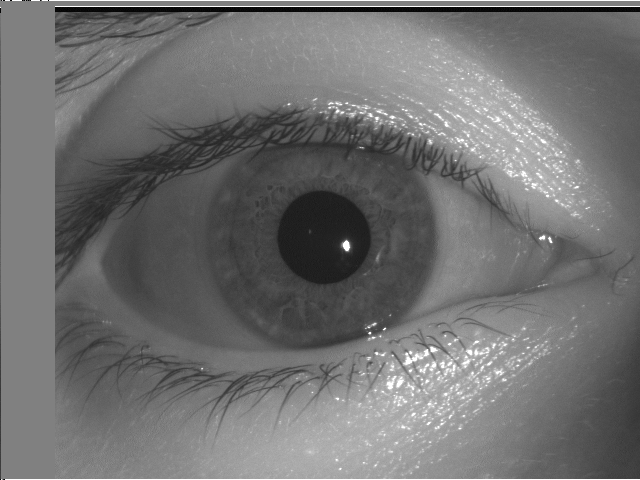}
        \caption{FN example (right)}
        \label{fig:test_example_fn_r}
    \end{subfigure} \\
    \caption{One common false positive (FP) iris pair and one common false negative (FN) iris pair made by all three models.}
    \label{fig:test_example}
\end{figure}

Table \ref{tab:test-res} presents the test outcomes for all three input types over five experimental runs. The model using original iris images achieves the best overall performance, exceeding the other two in terms of accuracy, recall, and F1 score. Although the non-iris-only input obtains the highest precision score, the advantage is marginal. Fig. \ref{fig:test_example} shows examples of false positive and false negative predictions for all three models.

Each input type maintains a clear separation between MZ and NMZ pairs, as demonstrated by the lower average distance for positive pairs and the higher average distance for negative pairs. However, the separation is not as optimal as observed during training. Notably, the non-iris-only input displays a greater separation for the negative pairs compared to the other two input types.

\section{DISCUSSION} \label{sec:discussion}

\subsection{Results Analysis}

The testing results indicate that the original input model, which utilizes the complete eye region, outperforms both the iris-only (masked) and the non-iris-only (inversely masked) models. At first, we anticipated the masked input model to excel by focusing exclusively on the iris's unique biometric features. However, these results suggest that non-iris ocular features or contextual information present in the original images contribute positively to distinguishing MZ from NMZ pairs, at least under the given training conditions.

Why might this be the case? Several hypotheses can be proposed, though drawing definitive conclusions would require further controlled experiments and detailed ablation studies:

\begin{itemize}
    \item \textbf{Contextual Cues:} The additional features surrounding the iris—such as sclera patterns, eyelid contours, or subtle lighting variations—help the model form more robust embeddings. Even though these cues are not traditionally considered primary biometric traits, they offer supplemental distinguishing information.
    \item \textbf{Regularization Effects:} The presence of extraneous features in the original input might act as a form of regularization, preventing the model from over-fitting to extremely subtle iris patterns. In contrast, the iris-only model may focus too narrowly on features that are difficult to generalize.
    \item \textbf{Segmented Iris Variability:} The segmentation and masking process may introduce noise or remove slight but critical details. If the masked images lose some subtle iris boundary information or produce less reliable pixel-level consistency, the resulting embeddings might be less discriminative.
\end{itemize}

As for the model with non-iris-only input, while it does not outperform the model with original input, its reasonable performance suggests that non-iris regions do aid in classification. It is surprising that non-iris-only input outperforms iris-only input. The numbers clearly indicate that the non-iris region provides valuable information for discrimination. It challenges our initial expectations about the primary location of critical signals.

\subsection{Pupil Dilation Effects}

\begin{figure}
    \centering
    \includegraphics[width=\linewidth]{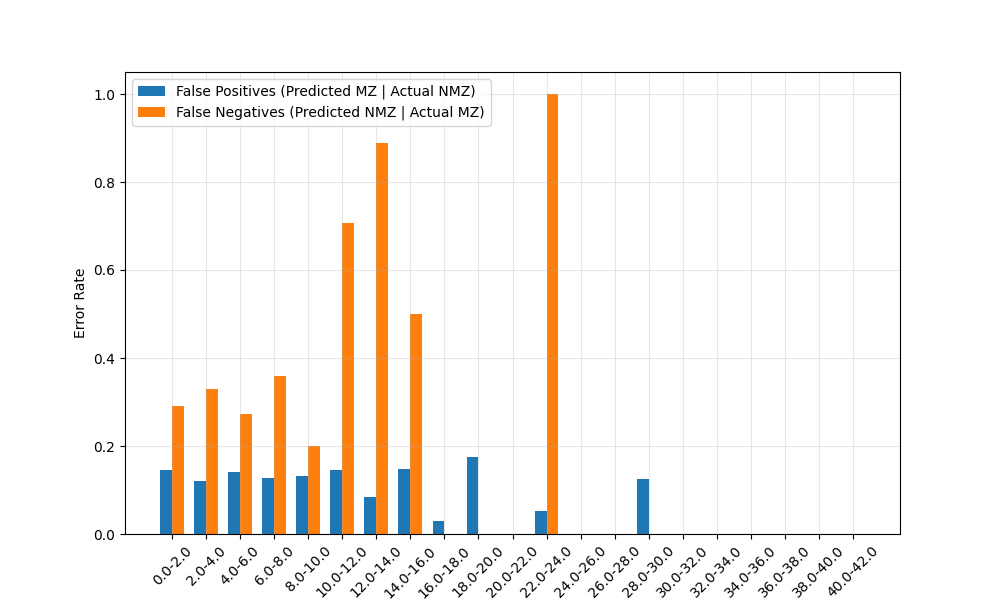}
    \caption{False positive rates and false negative rates vs. pupil-iris ratio differences between test pairs.}
    \label{fig:error-rates}
\end{figure}

In addition to the type of input image, the consistency of pupil dilation between paired iris images appears to be a relevant factor influencing performance, particularly for MZ pair detection.  Further analysis of error patterns reveals that the false negative rate for MZ pairs increases as the difference in pupil-iris ratio between the two images in a pair becomes larger (see Fig. \ref{fig:error-rates}). This observation suggests that when the pupil dilation differs significantly between two iris images from MZ twins, the model is more likely to incorrectly classify them as non-monozygotic. Conversely, the false positive rate for NMZ pairs is less affected by differences in pupil-iris ratio.  This indicates that controlling or normalizing pupil dilation could be beneficial for improving the accuracy of MZ iris pair detection. While not explicitly addressed in the current model, future work could investigate methods to control or normalize pupil dilation during image acquisition to potentially decrease false negative errors and enhance the overall robustness of the system.

\subsection{Comparison with prior work on human performance}
\label{sec:comp-human-perf}

Hollingsworth et al. conducted a study on the accuracy of human viewers in classifying iris image pairs as either matching (MZ) or non-matching (NMZ) \cite{hollingsworth2011genetically}. In their experiments, they compared human performance when viewing three different types of image inputs: whole iris images, iris texture alone, and the periocular region alone. Their findings revealed that human viewers achieved the highest accuracy when evaluating whole iris images, aligning closely with the results we observed using a ResNet-based model. Interestingly, they also found that human viewers performed better when viewing only the iris texture compared to the periocular region alone. This result contrasts with our findings. This discrepancy suggests the intriguing possibility that deep learning models may be learning to extract more useful and discriminative features from the periocular region than humans are able to. Hollingsworth et al. reported that human viewers achieved an accuracy of just over 80\% when classifying MZ and NMZ iris image pairs from identical twins. This performance level is comparable to the accuracy achieved by our model, further highlighting the parallels and differences in feature utilization between human perception and deep learning models.

Ultimately, these findings emphasize the complexity of iris biometrics when dealing with MZ and NMZ pairs. Future work might explore refined segmentation techniques, attention mechanisms to highlight relevant regions, or more sophisticated feature extraction strategies. Such efforts could clarify why contextual information matters and how best to leverage it for robust biometric verification in genetically related populations.

\section{FUTURE WORKS}

The results of our study highlight several avenues for future exploration and improvement. Although the original input model demonstrated the strongest overall performance, further investigations can help clarify the factors influencing these outcomes and advance the field of iris-based MZ/NMZ separation.

One intuitive way to improve the model performance is expanding the natural MZ iris dataset size. In Sec. \ref{sec:dataset-construction}, we mentioned that the MZ iris dataset is not sufficient for our experiment. Since iris texture can vary significantly across different populations and may be affected by ocular conditions, enlarging the dataset to encompass a broader range of demographic groups, iris pathologies, and capture conditions could help the network better model subtle intra-class variations and reduce biases introduced by limited data. Larger datasets might also facilitate more rigorous statistical evaluations of model performance, providing deeper insights into iris biometrics.

Future work could employ more sophisticated segmentation models to isolate the iris with greater precision. Improved segmentation could reduce artifacts introduced during the masking process, potentially allowing the iris-only model to better leverage inherent textural cues. Iris image normalization is often used to account for variations in scale, rotation, and illumination. This might also enhance the model's ability to focus on intrinsic features.

We notice that the ISO standard quality check implementation still has room for improvement. Although the assembled dataset filters out most of the low-quality iris images, there still exist outliers with uncentered gaze angle and unexpected blurriness. Further improving the quality check could control the input data better and introduce less noise.

Integrating spatial or channel-wise attention \cite{woo2018cbam} could help the model learn to focus on the most discriminative iris features without being distracted by irrelevant areas. Additionally, techniques such as Grad-CAM \cite{selvaraju2017grad} can be used to visualize and understand which regions of the image drive the final decision, offering insights into whether contextual cues or iris-boundary details influence performance. However, this technique is mostly used with neural networks for classification tasks rather than contrastive learning, requiring further investigation to adapt Grad-CAM to this method. Similarly, future work could leverage RISE \cite{petsiuk2018rise}, a technique for explaining black-box models by generating saliency maps. RISE could provide a more systematic understanding of which regions of the iris image contribute most significantly to model decisions. Combining Grad-CAM and RISE could help identify and mitigate potential biases in the learned representations while offering complementary insights into the model's interpretability and focus.

Exploring transfer learning between different datasets or employing domain adaptation techniques could enhance robustness. Models trained on one dataset might generalize more effectively to another after fine-tuning, potentially uncovering universally discriminative iris features for MZ/NMZ separation.

Further studies may consider collecting a more robust dataset with more temporal information. The twin iris images, especially those in the testing dataset, are mostly taken on the same day. Although multiple images are taken, the subjects are in a similar condition. For example, some subjects have mascara on, and it could become a visual clue for the model to deviate its focus.

Systematic ablation studies examining which layers and filters contribute most to MZ/NMZ discrimination would clarify the feature hierarchy within the network. Such studies might inform the design of custom architectures or loss functions optimized for subtle biometric distinctions. Our study uses ResNet-18 as the backbone of the recognition system. Future work could explore whether increasing the depth or complexity of the network architecture improves performance in separating MZ and NMZ iris. For example, architectures such as vision transformers (ViTs) \cite{dosovitskiy2020vit}.

Investigating the impact of different similarity measures, such as Euclidean distance and cosine similarity, on the performance of contrastive learning-based models could provide deeper insights. While cosine similarity focuses on the angle between feature vectors, Euclidean distance captures absolute differences. A comparative study might reveal which metric better captures the differences and similarities required for MZ/NMZ separation.

By pursuing these directions, future work can improve segmentation fidelity, interpretability, and generalizability, ultimately leading to more reliable and explainable solutions for accurately grouping MZ pairs and separating them from NMZ counterparts in iris recognition tasks.

\section{CONCLUSION}

In this study, we present the first automated approach to address the challenge of identifying MZ iris pairs from NMZ pairs — a problem that traditional iris recognition systems struggle with, given the subtle genetic similarities among MZ twins. By leveraging a Siamese network architecture and contrastive learning, we learned compact, discriminative representations capable of capturing nuanced patterns that define genetic similarity in the iris.

Our examination of different input conditions — original images, iris-only images, and non-iris-only images — revealed that using the complete iris image yields the best performance. Interestingly, the non-iris region alone demonstrated better accuracy than the iris-only region. This finding challenges the notion of the iris region being the sole critical component for distinguishing MZ and NMZ iris pairs. The understanding of the interplay between different iris image regions offered new insights into the underlying biometric signals.

The results demonstrate that it is indeed feasible to accurately cluster MZ pairs while isolating NMZ pairs, thereby potentially serving applications in biometric verification, security, and forensics. Future work may focus on refining segmentation techniques, improving interpretability through attention-based models, and extending the approach to other challenging biometric modalities. Ultimately, this research provides a robust foundation for more refined and trustworthy iris recognition systems that better understand and leverage the complex interplay between genetic similarity and individual uniqueness.

\section{ACKNOWLEDGMENT} \label{sec:ack}

We dedicate this paper to the memory of John Daugman, the father of the field of iris biometrics, who passed away in 2024. This work builds on the efforts of many current and past researchers in the Computer Vision Research Lab at Notre Dame; in particular, we thank Meghan McClain for her recent work.

In Fig. \ref{fig:teaser}, the top row is a pair of iris images from monozygotic twins.

{\small
\bibliographystyle{ieee}
\bibliography{main}
}

\clearpage
\appendix

\subsection{ISO/IEC 29794-6 Quality Metrics}
\label{appx:iso-metrics}

Here are all the computed quality metrics as mentioned in Sec. \ref{sec:iso-metrics}:

\begin{itemize}
    \item Usable iris area: Assesses whether a sufficient portion of the iris region is visible and is not occluded by eyelids, eyelashes, or reflections.
    \item Iris-sclera contrast: Measures the contrast between the iris and the sclera for clear boundary detection.
    \item Iris-pupil contrast: Evaluates the contrast between the iris and the pupil for accurate segmentation.
    \item Pupil boundary circularity: Assesses how closely the shape of the pupil approximates a perfect circle, indicating proper image capture without distortion.
    \item Grayscale utilization: Analyzes the distribution of grayscale values to confirm that the image has sufficient dynamic range, avoiding under- or over-exposed.
    \item Iris radius: Checks that the iris is captured at an adequate size within the image.
    \item Pupil-iris ratio: Computes the ratio of pupil diameter to iris diameter, filtering out images with abnormal pupil dilation.
    \item Iris-pupil concentricity: Measures the alignment of the pupil and iris centers to detect off-axis gaze or misalignment during capture.
    \item Margin adequacy: Ensures that the iris region is not too close to the image borders, preventing loss of critical information.
    \item Sharpness: Assesses the image focus to ensure that iris texture details are preserved.
    \item Motion blur: Detects blurriness caused by either subject movement or camera shaking.
\end{itemize}

The most important metric for our study is the pupil-iris ratio, since it significantly affects the iris texture.

\subsection{Ablation Studies}

We conducted several additional experiments to explore the factors that could potentially affect the experimental results.

\subsubsection{Model Backbone Depth}
\label{appx:abl-backbone-depth}

In the original experiment design, we only used the ResNet-18 architecture as our model backbone. There are more complex ResNet architectures, including ResNet-34, ResNet-50, and ResNet-101 \cite{he2016resnet}. In this section, we conducted the experiments under the same setup but with these 3 different ResNet architectures as backbones. Table \ref{tab:backbone-res} summarizes the testing accuracy and F1 scores, measuring the overall performance. The results indicate that ResNet-50 marginally outperformed other variants, while ResNet-101 exhibited comparatively lower metrics. ResNet-18 and ResNet-34 demonstrated similar performance, with overlapping uncertainty ranges across both evaluation measures. Overall, the four models do not show significant differences with each other, suggesting that increasing backbone complexity does not consistently improve performance for this task. We suspect that it might require a different architecture to capture the subtle visual clues that the ResNet-based backbones are unable to detect in the iris images and achieve over 0.9 accuracy.

\subsubsection{Distance Threshold}
\label{appx:abl-dist-thres}

In Sec. \ref{sec:loss}, the distance threshold was initially set to 0.5 based on its position as the midpoint between positive and negative margins. While training and validation distances demonstrated ideal convergence, test-phase positive and negative distances showed divergence. To investigate threshold sensitivity, we evaluated values of 0.2, 0.4, 0.6, and 0.8 using the original-input ResNet-18 architecture, conducting five independent trials per configuration.  
Experimental outcomes revealed identical mean distances and matching standard deviations across all threshold values compared to the 0.5 baseline. This consistency across threshold variations suggests the parameter has no measurable impact on distance distributions in this experimental setup.

\begin{table}[t]
\centering
\begin{tabular}{ccc}
\Xhline{1pt}
\textbf{Backbone} & \textbf{Accuracy} & \textbf{F1 Score} \\
\hline
ResNet-18 & $0.81\pm0.024$ & $0.81\pm0.034$ \\
ResNet-34 & $0.81\pm0.078$ & $0.80\pm0.012$ \\
ResNet-50 & $0.84\pm0.023$ & $0.83\pm0.026$ \\
ResNet-101 & $0.80\pm0.031$ & $0.78\pm0.048$ \\
\Xhline{1pt}
\end{tabular}
\caption{Testing results of different ResNet-based backbones.}
\label{tab:backbone-res}
\end{table}

\end{document}